\documentclass[11pt]{article}

\usepackage{acl}
\usepackage{multirow}
\usepackage{times}
\usepackage{latexsym}
\usepackage[T1]{fontenc}
\usepackage[utf8]{inputenc}
\usepackage{inconsolata}
\usepackage{graphicx}
\usepackage{amsmath}
\usepackage{booktabs}
\usepackage{amssymb}
\usepackage[table]{xcolor}
\definecolor{dropred}{RGB}{255,235,235}
\usepackage{xcolor}
\usepackage{booktabs}
\usepackage{float}
\usepackage[most]{tcolorbox}
\definecolor{lightblue}{RGB}{220,235,250}
\usepackage{colortbl}
\usepackage{makecell}
\newcommand{\drop}{\textcolor{red}{$\downarrow$}}
\newcommand{\gain}[1]{\textcolor{green!50!black}{\scriptsize +#1}}

\title{Rethinking Continual Experience Internalization for\\
Self-Evolving LLM Agents}

\author{
\normalfont
Jingwen Chen$^{1}$\thanks{Equal contribution.} \quad
Wenkai Yang$^{1}$\footnotemark[1] \quad
Shengda Fan$^{1}$ \quad
Wenbo Nie$^{2}$ \quad
Chenxing Sun$^{3}$ \\
Shaodong Zheng$^{3}$ \quad
Yangen Hu$^{3}$ \quad
Lu Pan$^{3}$ \quad
Ke Zeng$^{3}$ \quad
Yankai Lin$^{1}$\thanks{Corresponding author.} \\
$^{1}$Gaoling School of Artificial Intelligence, Renmin University of China \\
$^{2}$School of Software, Beihang University \\
$^{3}$Meituan \\
\texttt{cjw259wen@outlook.com} \quad
\texttt{yankailin@ruc.edu.cn}
}

\begin{document}
\maketitle
\begin{abstract}
Experience internalization converts contextual experience from past interactions into reusable parametric capability, offering a promising path toward continual learning in large language models (LLMs). While prior work has predominantly focused on single-iteration transfer, we discover that under multi-iteration experience learning, existing methods suffer from a progressive capability collapse rather than compounding improvement. We systematically examine this failure through three vital dimensions of experience internalization:
(1) \textbf{\emph{Experience Granularity}}: We find that principle-level experience is more durable than instance-level experience, as it effectively abstracts transferable strategies away from trajectory-specific details.
(2) \textbf{\emph{Experience Injection Pattern}}: Our analysis reveals that step-wise injection significantly outperforms global injection by aligning experience with intermediate decision states, a property that is critical for long-horizon tool use.
(3) \textbf{\emph{Internalization Regime}}: We demonstrate that off-policy context-distillation on high-quality teacher trajectories provides a substantially more stable training signal than on-policy context-distillation, which is inherently limited by local corrections on student-induced flawed states.
Together, these insights yield a simple yet robust recipe for stable and sustainable experience internalization, providing concrete guidance for engineering self-evolving and continually learning LLMs. The code and data for this work are available at \url{https://github.com/RUCBM/ExpInternalization}.
\end{abstract}

\section{Introduction}
\label{sec:intro}
\begin{figure}[t]
    \centering
    \includegraphics[width=\linewidth]{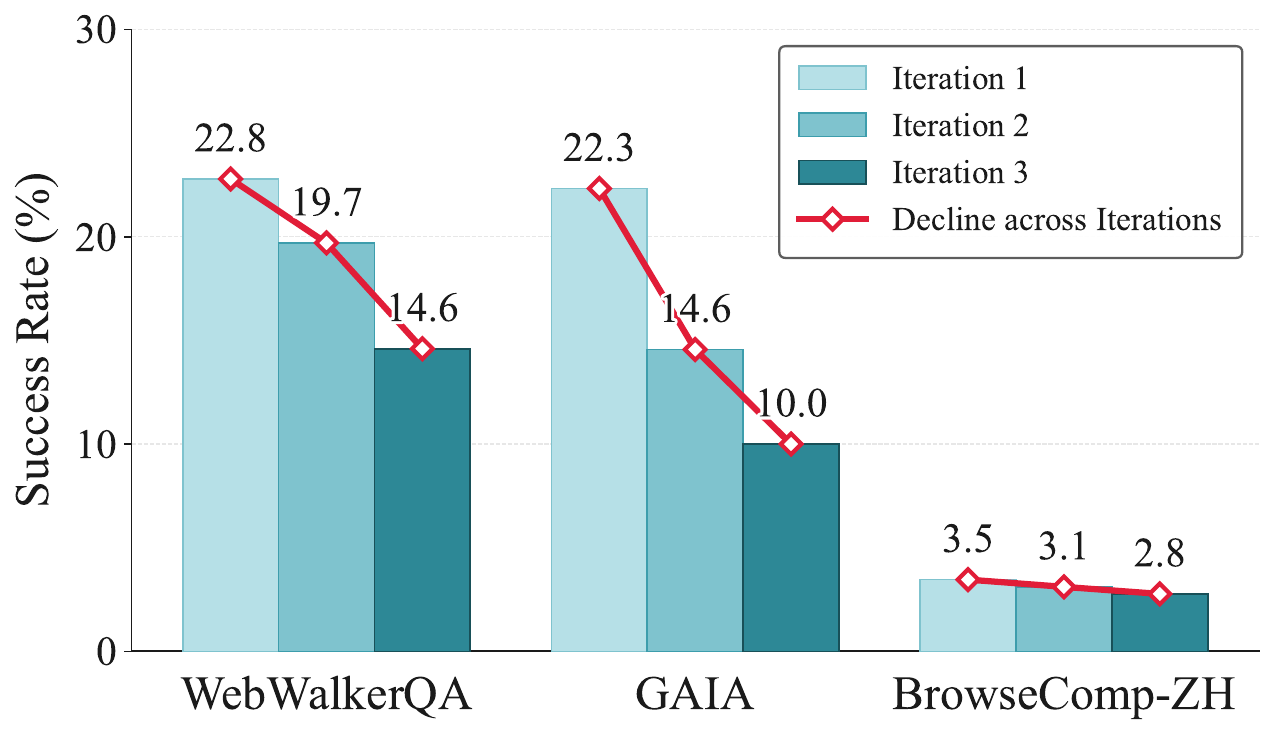}
    \caption{Performance degradation under iterative on-policy context-distillation.}
    \label{fig:preliminary}
    \vskip -0.05in
\end{figure}

The capability for continual learning \citep{wu2024continual,gao2025survey,wang2023voyager} is essential for building autonomous and adaptive LLM agents. Toward this end, learning from experience~\citep{zhao2024expel,shinn2023reflexion,silver2025welcome} offers a promising path, enabling LLMs to acquire generalizable knowledge from past interactions and continuously improve through future interactions. 
In-context learning (ICL) \citep{dong2024survey,brown2020language} represents the most direct exploitation of experience by presenting it to the model as context. However, this paradigm is bounded by in-context capacity and prone to context collapse \citep{ace2025} as the experience pool grows.
This motivates experience internalization~\citep{snell2022learning,deng2024explicit,opcd2026,kujanpaa2024efficient,charakorn2026doc}, which converts context-dependent experience use into parametric capability.
Most recent work on experience internalization adopts on-policy context-distillation~\citep{opcd2026,ye2026online,shenfeld2026self} and achieves strong performance in a single iteration of internalization. 
However, existing approaches largely overlook the necessity of iterative experience internalization, which is a cornerstone of the continual learning paradigm.
Through a preliminary study, we reveal a critical vulnerability: as shown in Figure~\ref{fig:preliminary}, \textbf{current methods fail to sustain this self-evolving process, with performance collapsing as self-evolution proceeds}.

In this study, we rethink why current experience internalization paradigms fail under multi-iteration experience learning. 
We attribute these failures to three stages of the transfer: how experience is represented, how it shapes teacher supervision, and which trajectory distribution is used to transfer the resulting behavior into the student.

First, for \textbf{\emph{Experience Granularity}}, we find that principle-level experience is more suitable for internalization than instance-level experience. 
By abstracting transferable strategies and failure patterns from trajectory-specific details, principle-level experience provides a more generalizable signal and reduces the risk of reinforcing instance-specific behaviors across iterations.
In addition to experience granularity, we further explore the effect of \textbf{\emph{Experience Injection Pattern}}.
We find that step-wise injection outperforms global injection by aligning relevant experience with intermediate decision states.
This state-aligned use of experience is especially important in long-horizon tool-use tasks, where global injection can fail to preserve the model's ability to use newly generated experience in later self-evolution iterations.
However, degradation can still occur under principle-level experience and step-wise injection, motivating us to examine \textbf{\emph{Internalization Regime}}, which specifies the trajectory distribution for transferring experience-conditioned behavior.
We find that on-policy context-distillation delivers strong gains in a single iteration but fails to sustain them across multiple iterations. Since supervision is built on student-induced trajectories, the teacher is reduced to local corrections on flawed states, rather than coherent demonstrations of experience-guided behavior. Off-policy context-distillation, by contrast, trains on high-quality teacher-generated trajectories, providing a more stable signal for experience internalization and self-evolution.
Overall, we systematically study experience internalization across these three dimensions and propose a simple recipe for sustainable internalization. 
These findings provide practical guidance for designing LLM agents that can sustain experience-based self-evolution across iterations.

\section{Related Work}
\label{sec:related}
\subsection{Learning from Experience}

\paragraph{Context-Based Experience Learning}
The experience accumulated from the interaction trajectories of LLM agents provides a valuable resource for improving agent behavior. 
Recent work reuses such experience as contextual guidance without parameter updates.
These methods can be broadly organized into storage, reflection, and abstraction \citep{memorysurvey2026}: preserving trajectories for retrieval \citep{zheng2024synapse}, refining stored experience through self-feedback \citep{shinn2023reflexion,xu2026mem}, and generalizing experience into reusable forms such as skills, strategies, or summarized experiential knowledge \citep{fan2026generalizing,ace2025,cai2025tfgrpo}.
However, context-based methods retain experience as inference-time context, leaving their benefits bounded by the model's in-context learning ability and vulnerable to context collapse when experience accumulates~\citep{ace2025}. This motivates our study of sustainable experience internalization beyond inference-time context.

\paragraph{Experience Internalization}

Context distillation~\citep{askell2021general,snell2022learning} provides a way to internalize experience into model parameters by aligning an experience-free student with an experience-aware teacher.
Early formulations are often off-policy~\citep{hinton2015distilling,yang2025distilling}, where the student is trained on teacher-generated trajectories but may suffer from training--inference mismatch~\citep{agarwal2024gkd}.
Recent work has therefore shifted toward on-policy context distillation~\citep{gu2024minillm,opcd2026,zhao2026self,yang2026learning,hou2026uni,fu2026revisiting,rethinking_opd}, which supervises trajectories sampled from the student to improve distributional consistency.
However, existing works focus on single-round transfer, leaving the stability of multi-iteration internalization underexplored. We address this gap by studying sustainable experience internalization across self-evolution cycles.


\subsection{Self-Evolving LLM Agents}
Self-evolving LLM agents refer to agent systems that iteratively improve their behavior by leveraging interaction data, feedback signals, and self-generated experience \citep{tao2024survey,fang2025comprehensive}.
Existing work has explored self-evolution at both the policy and component levels. Policy-level methods \citep{huang2025r,zhao2026absolute, fan2026darc} update the agent model from interaction trajectories and feedback, whereas component-level methods \citep{xu2026mem,liu2025contextual} evolve external structures such as memory, tools, skills, or experience libraries.
Recent work further couples model training with experience evolution in a closed loop \citep{xia2025agent0,ye2026online}, iteratively training from the experience pool and refreshing it with trajectories from the updated model.
Effective experience-based self-evolution requires experience evolution and model improvement to reinforce each other across rounds.
We therefore study how experience representation and internalization can strengthen this loop and support subsequent policy improvement.


\section{Formulation}
\label{sec:formulation}

We formalize continual experience internalization and introduce the notation used in our analysis.

\paragraph{Agent Trajectories and Experience Pool.}
Following ReAct~\citep{yao2022react}, an agent policy $\pi_\theta$ interacts with an environment through interleaved reasoning and action steps, where $\mathcal{A}$ denotes the action space.
Given a user query $x$, at each step $t$, the agent generates a thought $\tau_t$ and an action $a_t \in \mathcal{A}$ conditioned on the history $\mathcal{H}_{t-1}$, where $a_t$ is either a tool call or a terminal answer.
Tool calls return observations $o_t$, forming a trajectory 
$\mathcal{H}_T = \big(x,(\tau_1,a_1,o_1),\ldots,(\tau_T,a_T,o_T)\big)$ 
evaluated by a task-level reward $r(\mathcal{H}_T)$.
Following prior work on experience extraction~\citep{cai2025tfgrpo}, we summarize trajectories into natural-language experience with DeepSeek-V4~\cite{deepseekai2026deepseekv4} unless otherwise specified, and denote the resulting pool as $\mathcal{E}=\{e_1,\ldots,e_N\}$.

\paragraph{Experience Distillation.}
Experience internalization distills an experience-aware teacher $\pi_T$ into an experience-free student $\pi_\theta$.
The teacher can access injected experience $\mathcal{E}_t \subseteq \mathcal{E}$ during supervision construction, while the student acts without experience at deployment.
For brevity, let $h_{t-1}=\mathcal{H}_{t-1}$, 
$p_t=\pi_T(\cdot \mid h_{t-1},\mathcal{E}_t)$, and 
$q_t=\pi_\theta(\cdot \mid h_{t-1})$.
We consider two internalization regimes.
In \textbf{off-policy context-distillation}, trajectories are generated by the teacher and the student matches the teacher distribution with forward KL:
\begin{equation}
\mathcal{L}_{\mathrm{off}}(\theta)
=
\mathbb{E}_{\mathcal{H}\sim \pi_T}
\sum_{t=1}^{T}
D_{\mathrm{KL}}\!\left(p_t \,\Vert\, q_t\right).
\end{equation}
In \textbf{on-policy context-distillation}, trajectories are generated by the student and the teacher supervises student-induced states with reverse KL:
\begin{equation}
\mathcal{L}_{\mathrm{on}}(\theta)
=
\mathbb{E}_{\mathcal{H}\sim \pi_\theta}
\sum_{t=1}^{T}
D_{\mathrm{KL}}\!\left(q_t \,\Vert\, p_t\right).
\end{equation}

\begin{figure*}
    \centering
    \includegraphics[width=\textwidth]{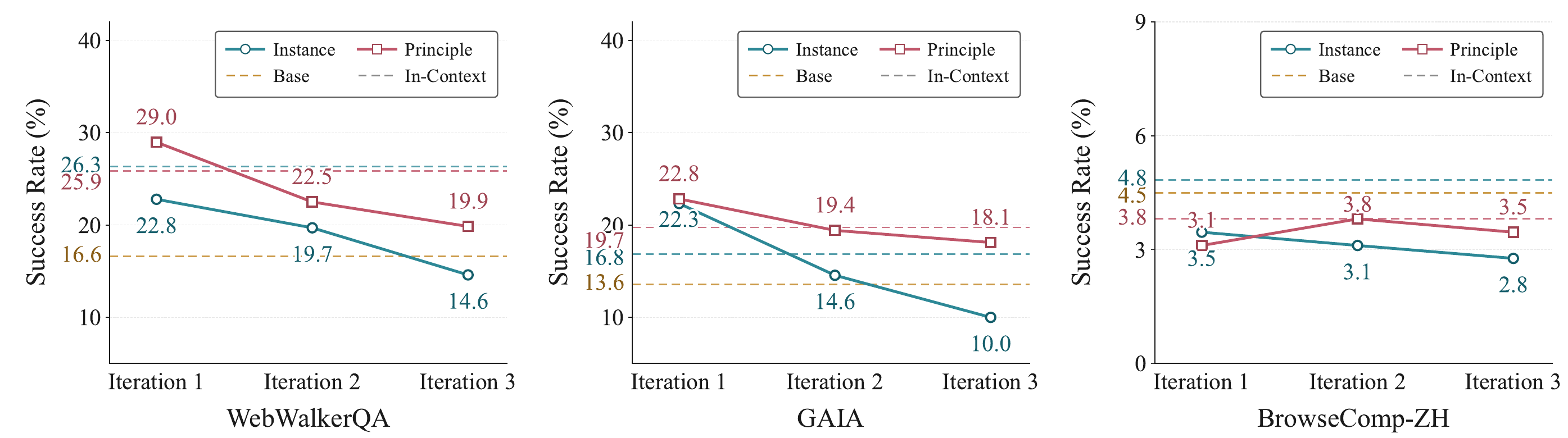}
    \caption{
Effect of Experience Granularity on Qwen3-4B-Instruct-2507 under iterative on-policy context-distillation.
Dashed lines denote base and in-context performance.
}
    \label{fig:granularity}
\end{figure*}

\paragraph{Continual Experience Internalization.}
To study experience internalization beyond a single update, we consider an iterative process indexed by $k=0,1,\ldots,K$.
At iteration $k$, the current policy $\pi_{\theta^{(k)}}$ interacts with the environment and produces trajectories $\mathcal{D}^{(k)}=\{\mathcal{H}^{(k)}_i\}$.
These trajectories are summarized into an experience pool $\mathcal{E}^{(k)}$.
The same policy, when conditioned on $\mathcal{E}^{(k)}$, serves as an experience-aware teacher for training the next experience-free student $\pi_{\theta^{(k+1)}}$:
\begin{equation}
\theta^{(k+1)}
=
\operatorname{Internalize}
\big(
\theta^{(k)}, \mathcal{E}^{(k)}
\big).
\end{equation}
This closed loop captures the promise of continual experience learning: an agent may transform accumulated experience into reusable capability as its policy evolves.
Therefore, experience internalization should be evaluated not only by single-iteration gains, but also by whether such gains can be sustained across iterations.

\paragraph{Dimensions of Experience Internalization.}
In this framework, we study three dimensions that shape sustained experience internalization.
\textbf{\emph{Experience Granularity}} specifies the abstraction level of the experience pool $\mathcal{E}^{(k)}$.
Instance-level experience preserves trajectory-specific details, while principle-level experience abstracts reusable strategies, decision rules, and failure patterns.
\textbf{\emph{Experience Injection Pattern}} specifies how experience is provided to the teacher during supervision construction.
Under global injection, the teacher uses a fixed experience context $c^{\mathrm{glob}}=[x;\mathcal{E}^{(k)}]$ for the whole trajectory, inducing the teacher distribution $p_t^{\mathrm{glob}}=\pi_T(\cdot \mid h_{t-1}, c^{\mathrm{glob}})$.
Under step-wise injection, an LLM-based selector $R_\phi$ selects experience according to the current interaction history, $\mathcal{E}^{\mathrm{step}}_t=R_\phi(h_{t-1},\mathcal{E}^{(k)})$, inducing $p_t^{\mathrm{step}}=\pi_T(\cdot \mid h_{t-1},\mathcal{E}^{\mathrm{step}}_t)$.
\textbf{\emph{Internalization Regime}} specifies the trajectory distribution on which experience-conditioned teacher behavior is transferred to the student, contrasting off-policy internalization on teacher-generated trajectories with on-policy internalization on student-induced trajectories.
Together, these dimensions define the design space for continual experience internalization in this work.

\section{Experimental Setup}
\label{sec:exp}

\paragraph{Models and Environment.}
We use Qwen3-4B-Instruct-2507 and Qwen3-8B~\citep{yang2025qwen3} as student models, with thinking mode disabled for Qwen3-8B.
The agent follows the ReAct-style interaction format with five tools: \textit{Search}, \textit{Visit}, \textit{Python}, \textit{Scholar}, and \textit{File Parser}.

\paragraph{Training Data and Experience.}
We construct a 15K-example training corpus from five public web-reasoning QA datasets: WebWalkerQA-silver~\citep{wu2025webwalker}, DeepDive~\citep{lu2025deepdive}, WebShaper~\citep{tao2025webshaper}, WebDancer~\citep{wu2026webdancer}, and SailorFog-QA~\citep{li2025websailor}.
We use this corpus to generate agent trajectories, extract natural-language experience, and then use the resulting experience pools to construct experience-conditioned supervision under the internalization regimes defined in Section~\ref{sec:formulation}.

\paragraph{Benchmarks and Metrics.}
We evaluate on WebWalkerQA~\citep{wu2025webwalker}, GAIA-Text-103~\citep{mialon2024gaia}, and BrowseComp-ZH~\citep{zhou2025browsecomp}.
Since WebWalkerQA-silver is included in our training corpus, we treat WebWalkerQA as in-domain and the other two as out-of-domain benchmarks.
We report Pass@1 on WebWalkerQA and BrowseComp-ZH with one rollout per query, and average accuracy on GAIA-Text-103 over three rollouts.
For brevity, we refer to GAIA-Text-103 as GAIA in tables.

\paragraph{Training and Inference.}
All methods are implemented with verl~\citep{sheng2024hybridflow}.
We train students using a learning rate of $1\times10^{-5}$, a batch size of 128, and 5 epochs on $8\times$ NVIDIA A800 GPUs.
During inference, we use temperature 0.7, allow at most $T_{\max}=100$ interaction steps, and set the context window to 32,768 tokens.


\begin{figure*}[t]
    \centering
    \includegraphics[width=\textwidth]{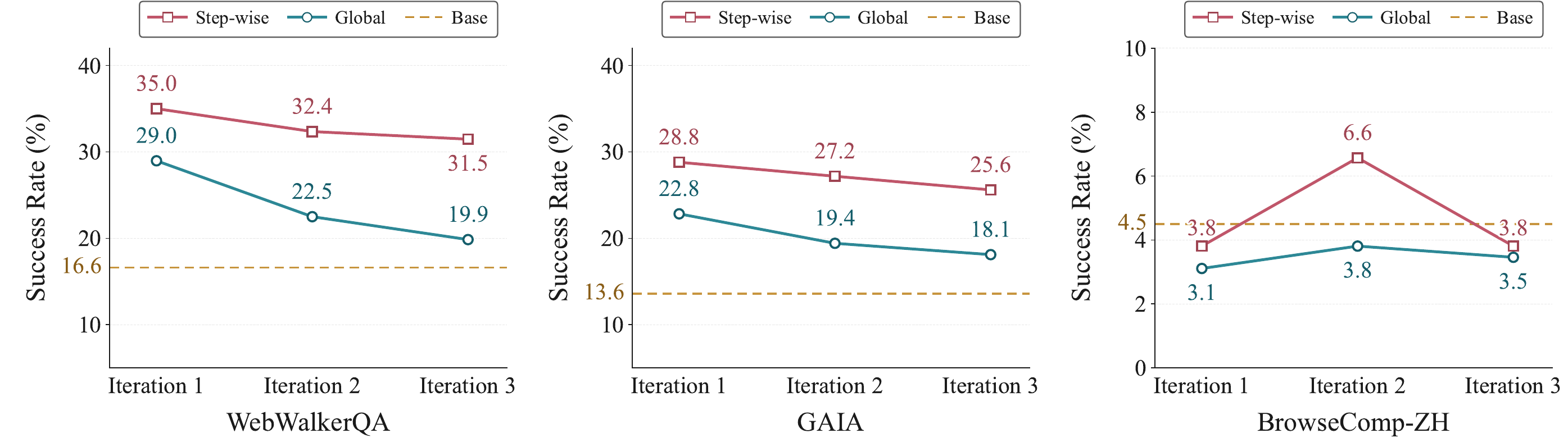}
    \caption{
Effect of Experience Injection Pattern on Qwen3-4B-Instruct-2507 under iterative on-policy context-distillation.
Dashed lines denote base performance.
}
    \label{fig:injection}
\end{figure*}

\section{Toward Stable Continual Experience Internalization}

\subsection{Effect of Experience Granularity}
\label{sec:granularity}
We first examine how \emph{Experience Granularity} shapes the reliability of experience internalization across iterations. We compare instance-level experience, which preserves trajectory-specific details, with principle-level experience, which abstracts reusable strategies, search principles, and failure patterns. Both are evaluated under in-context use and iterative internalization.

Figure~\ref{fig:granularity} shows that instance-level experience yields only transient gains. Although it improves performance in the first iteration, these gains quickly diminish as self-evolution proceeds and fall below the base model. This fragility stems from the localized content profile of instance-level data. In our sampled pool, 74.4\% of instance-level items contain specific URLs or domains, 57.3\% contain concrete numbers, and 93.9\% contain query- or entity-specific strings. Such trajectory-specific traces facilitate in-distribution exploitation but transfer poorly once the model encounters new queries or induces different trajectories.

Principle-level experience provides a durable signal by filtering out such local artifacts and retaining reusable decision rules.
In our sample, 84.0\% of principle-level items contain reusable strategy-like statements, compared with only 3.7\% of instance-level items.
This abstraction reduces dependence on source trajectories and better supports internalization across updated trajectory distributions.

Overall, instance-level experience mainly provides short-term gains, whereas principle-level experience offers a more stable basis for sustained multi-iteration self-evolution.

\subsection{Effect of Experience Injection Pattern}
\label{sec:injection}

Having established that principle-level experience provides a more suitable signal for internalization, we next examine how such experience should be injected into the teacher prompt when constructing supervision.
We fix the experience granularity to principle-level experience and study the two injection patterns under on-policy context-distillation, where trajectories are sampled from the student and the teacher supervises student-induced states.

Following Section~\ref{sec:formulation}, the two injection patterns induce different teacher distributions, $p_t^{\mathrm{glob}}$ and $p_t^{\mathrm{step}}$, while the student remains experience-free with $q_t=\pi_\theta(\cdot \mid h_{t-1})$.
Under on-policy distillation, both settings supervise the same student-induced trajectory distribution and differ only in the teacher distribution used as the distillation target.
The global-injection objective is therefore:
\begin{equation}
\mathcal{L}_{\mathrm{on}}^{\mathrm{glob}}(\theta)
=
\mathbb{E}_{\mathcal{H}\sim \pi_\theta}
\sum_{t=1}^{T}
D_{\mathrm{KL}}\!\left(q_t \,\Vert\, p_t^{\mathrm{glob}}\right).
\end{equation}

Here, the teacher uses a fixed trajectory-level experience context, whereas step-wise injection uses a state-dependent teacher distribution:
\begin{equation}
\mathcal{L}_{\mathrm{on}}^{\mathrm{step}}(\theta)
=
\mathbb{E}_{\mathcal{H}\sim \pi_\theta}
\sum_{t=1}^{T}
D_{\mathrm{KL}}\!\left(q_t \,\Vert\, p_t^{\mathrm{step}}\right).
\end{equation}


\subsubsection{Injection Pattern in Single-Iteration Internalization}

We first examine the single-iteration results in Figure~\ref{fig:injection}.
At Iteration~1, step-wise injection consistently yields stronger internalization than global injection.
This indicates that merely making experience accessible to the teacher is insufficient.
The injection pattern affects whether the experience can shape the teacher distribution used for distillation.


\definecolor{groupbg}{HTML}{E6EEF7}
\definecolor{gainGreen}{HTML}{1B7F3B}

\newsavebox{\gainbox}
\newcommand{\numgain}[2]{%
  \sbox{\gainbox}{\textcolor{gainGreen}{\textsubscript{\,+#2}}}%
  \hphantom{\usebox{\gainbox}}%
  #1%
  \makebox[0pt][l]{\usebox{\gainbox}}%
  \hphantom{\usebox{\gainbox}}%
}

\begin{table}[!t]
\centering
\small
\setlength{\tabcolsep}{3pt}
\renewcommand{\arraystretch}{1.3}

\begin{tabular}{l c c c}
\toprule
Injection & WebWalkerQA & GAIA & BrowseComp-ZH \\
\midrule
Global    & 23.2 & 16.8 & 4.5 \\
\rowcolor{groupbg}
Step-wise & \numgain{31.2}{8.0} & \numgain{22.7}{5.9} & \numgain{5.2}{0.7} \\
\bottomrule
\end{tabular}

\caption{Single-iteration effect of Experience Injection Pattern with Qwen self-generated experience.}
\label{tab:iteration1}
\end{table}

This result suggests that the utility of experience is determined not only by the experience pool itself, but also by whether its content is selected and injected at the appropriate supervision state.
Such state-specific selection is crucial in long-horizon tool-use tasks, because experience that helps search planning may become irrelevant, or even misleading, at later states where the model should verify evidence or decide whether to terminate.
Global injection treats experience as a fixed trajectory-level context, which can misalign the injected experience with the decision currently being supervised.
Step-wise injection mitigates this issue by selecting experience according to the current interaction history, turning experience from static background context into decision-relevant supervision.

This advantage is also evident when the experience is generated by the student-side model itself.
As shown in Table~\ref{tab:iteration1}, under the Qwen self-generated setting, step-wise injection improves over global injection across all three benchmarks, increasing WebWalkerQA from 23.2\% to 31.2\%.
Compared with using a stronger external model for experience extraction and selection, the Qwen self-generated setting relies on the student-side model itself, providing a more challenging test of whether the injection pattern can exploit weaker experience.
This indicates that step-wise injection can extract useful supervision from self-generated experience, supporting experience-based self-evolution.

\subsubsection{Injection Pattern in Iterative Internalization}

While single-iteration gains are valuable, the critical question for continual experience learning is whether an injection pattern can sustain improvement as the model and the experience pool co-evolve.
As shown in Figure~\ref{fig:injection}, global injection yields only transient improvements and degrades as self-evolution proceeds.
In contrast, step-wise injection maintains stronger performance across iterations, especially on WebWalkerQA and GAIA.
This indicates that experience injection pattern affects not only the current internalization step, but also the sustainability of experience internalization under iterative updates.

This distinction is particularly important under Qwen self-generated experience.
Since the experience pool is produced by the student-side model, it provides a more challenging source of supervision than experience generated by a stronger external model.
Figure~\ref{fig:qwen_self_evolution_experience_use} further shows that step-wise injection better preserves the model's ability to benefit from explicit experience across iterations.
After later internalization rounds, step-wise-trained models can still improve when the corresponding experience pool is provided in context, whereas global-injection models degrade in both in-context and internalized performance.
This indicates that step-wise injection helps the updated model continue to use its newly generated experience pool when serving as the teacher in later iterations.
Without it, the newly generated experience pool cannot provide effective supervision for subsequent internalization.
These results suggest that step-wise injection provides a viable path for experience-based self-evolution, while global injection fails to preserve the utility of experience as the model and experience pool co-evolve.

\begin{figure}[t]
    \centering
    \includegraphics[width=\linewidth]{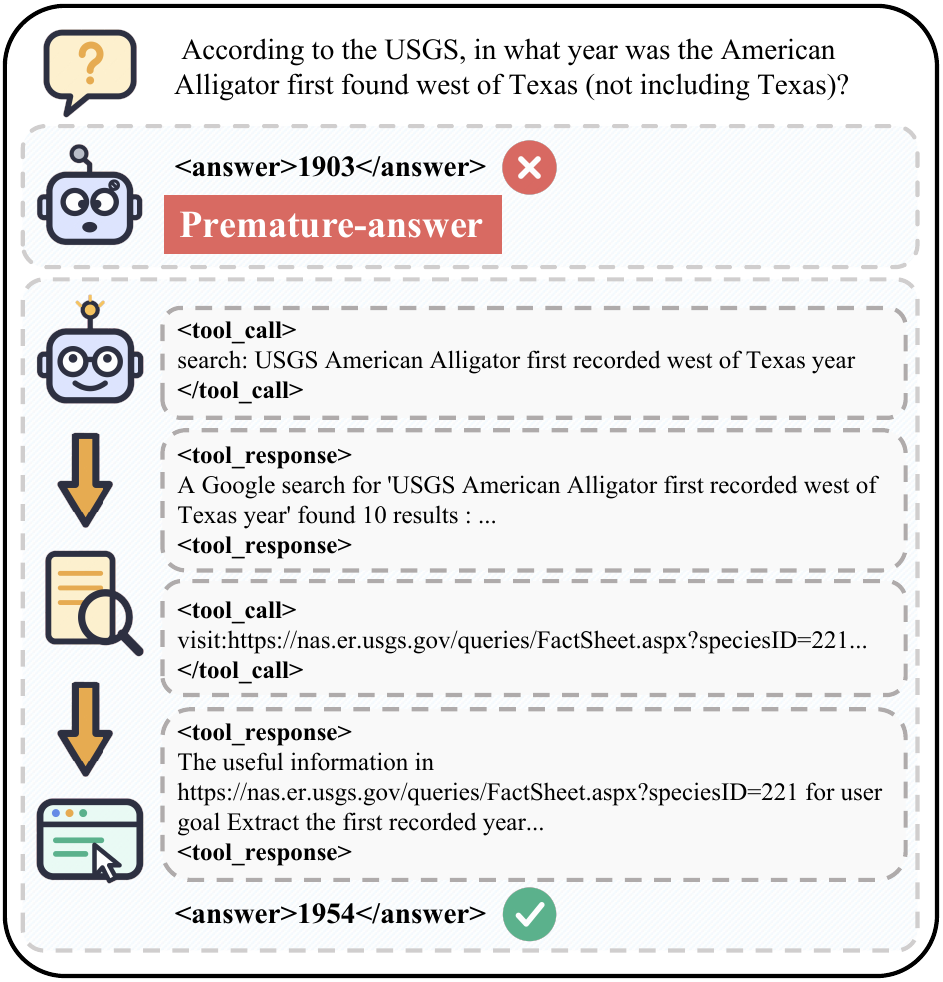}
    \caption{
Case study of \textbf{premature answering} under global injection.
After iterative training, the model trained with global injection terminates without invoking search tools, whereas step-wise injection preserves evidence-seeking tool use before answering.
}
    \label{fig:case}
\end{figure}

\begin{table}[!tbp]
\centering
\small
\setlength{\tabcolsep}{9pt}
\renewcommand{\arraystretch}{1.2}
\begin{tabular}{@{}lcc@{}}
\toprule
& \textbf{Global} & \textbf{Step-wise} \\
\midrule
Premature-answer rate & 63.82\% & 0\% \\
\bottomrule
\end{tabular}
\caption{
Premature-answer rate of third-iteration models under different injection patterns.
}
\label{tab:premature-answer}
\end{table}

\subsubsection{Why Step-wise Injection Supports Continual Experience Internalization}

We further analyze why step-wise injection better sustains continual experience internalization.
In iterative self-evolution, the model obtained from one internalization iteration is reused to construct supervision for the next.
Thus, the updated model must not only perform well without inference-time experience, but also retain \emph{experience-use ability}: the ability to further benefit from its corresponding experience pool at inference time, measured by the gap between in-context and experience-free inference.
This ability is necessary because the next-round teacher must use the updated experience pool to produce supervision.

\begin{figure*}
    \centering
    \includegraphics[width=\textwidth]{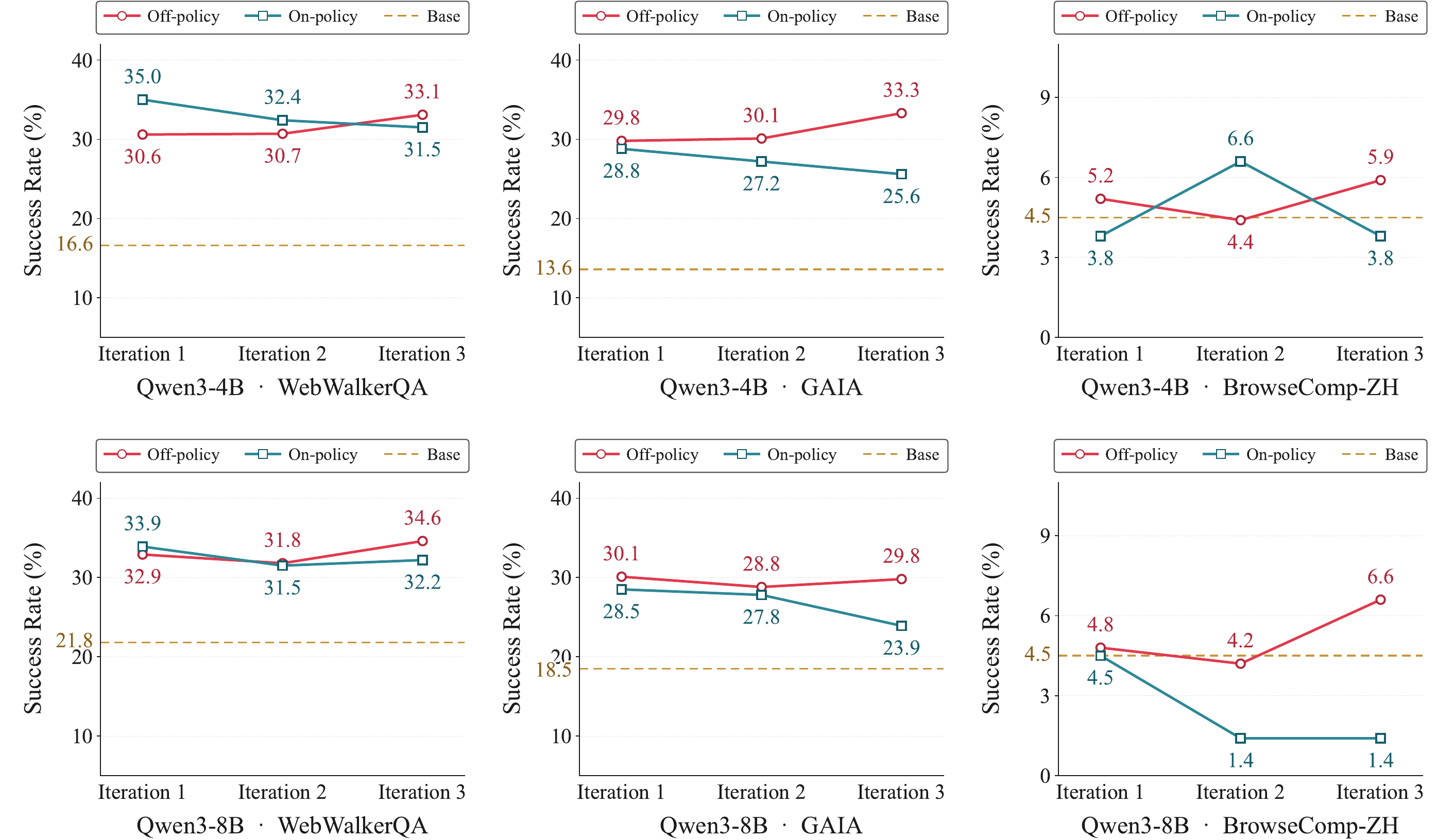}
    \caption{
Effect of Internalization Regime across self-evolution iterations.
We compare off-policy context-distillation with on-policy context-distillation under principle-level experience and step-wise injection on Qwen3-4B-Instruct-2507 and Qwen3-8B.
Dashed lines denote the base model without experience internalization.
}
    \label{fig:placeholder}
\end{figure*}

\begin{figure*}[t]
    \centering
    \includegraphics[width=\textwidth]{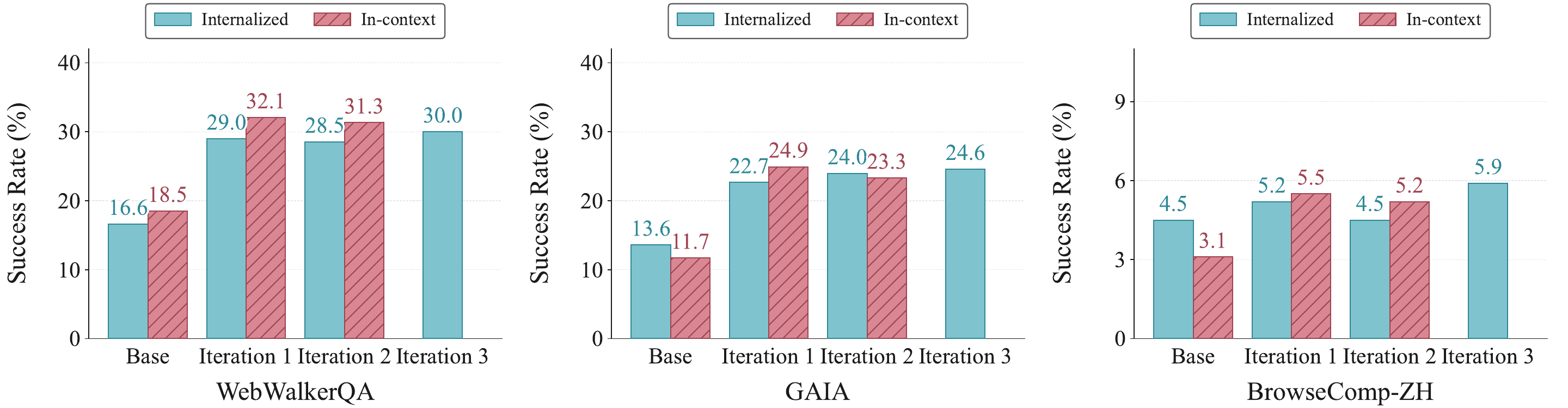}
    \caption{
Self-evolution performance of Qwen3-4B-Instruct-2507 under our final setting.
Cyan bars denote internalized inference without inference-time experience, while red bars denote in-context experience use with the corresponding experience pool.
The results show that our setting sustains performance gains across self-evolution iterations and preserves the model's ability to benefit from explicit experience.
}
    \label{fig:qwen_self_evolution_experience_use}
\end{figure*}

As shown in Figure~\ref{fig:qwen_self_evolution_experience_use} and Appendix Figure~\ref{fig:global_experience_use}, step-wise models continue to benefit from experience across iterations, whereas global-injection models degrade both with and without experience context.
This indicates that global injection not only fails to fully convert experience into parametric capability, but also weakens experience-use ability.
When reused in the next iteration, the model may provide weaker experience-conditioned supervision and destabilize the model--experience loop.


We also observe a premature-answer failure mode caused by the injection pattern.
As shown in Table~\ref{tab:premature-answer}, global injection directly produces an \texttt{\textless answer\textgreater} without any preceding \texttt{\textless tool\_call\textgreater} or tool observation in 63.82\% of the cases, while step-wise injection shows 0\%.
This failure stems not from the experience form itself, but from a mismatch between the injected experience and the current decision state.
Under global injection, the teacher receives the same fixed experience context throughout the whole trajectory, regardless of whether the current state requires search planning, evidence verification, or termination.
As a result, experience that is useful for later-stage decision making may be exposed too early, while experience relevant to the current state may not be emphasized.
This misalignment can shift the teacher distribution toward premature answer generation rather than continued tool use.
In contrast, step-wise injection selects experience according to the current interaction history, making the injected experience more decision-relevant at each state.
Figure~\ref{fig:case} illustrates this behavior: the global-injection model terminates before search, while the step-wise model continues evidence-seeking tool use.

Together, these analyses show that step-wise injection benefits both the current internalization round and the subsequent self-evolution loop.
By preserving experience-use ability and reducing exposure to irrelevant terminal information, it helps the internalized model remain an effective experience-aware teacher in later iterations, whereas global injection can weaken this role and make the model--experience loop less sustainable.

\subsection{Effect of Internalization Regime}
\label{sec:regime}
The previous two dimensions improve experience internalization, but performance can still degrade across self-evolution iterations.
We therefore revisit on-policy context-distillation, the dominant paradigm for experience internalization, and examine whether the transfer regime affects the stability of continual internalization.

\subsubsection{Trajectory Distribution and Supervision Coherence}
We compare on-policy context-distillation and off-policy internalization under the same principle-level, step-wise experience configuration, differing only in the trajectory distribution used for supervision. 
On-policy context-distillation samples trajectories from the current experience-free student and queries the experience-aware teacher on the resulting student-induced states. 
Off-policy internalization instead samples trajectories directly from the experience-aware teacher (i.e., the student conditioned on step-wise experience) and applies rejection sampling to retain successful trajectories.

This difference in trajectory distribution affects the coherence of the resulting supervision signal.

For on-policy context-distillation, supervision is fundamentally reactive.
Because the preceding trajectory is generated by the student without experience, the teacher can only provide corrections on states that may already be inefficient or off target.
When the student has deviated substantially from a useful search path, the teacher may struggle to provide valid guidance on these degraded states.
As a result, on-policy supervision can improve localized decisions, but it does not necessarily demonstrate how experience should guide a coherent trajectory.
This limitation is especially important in long-horizon tool use, where search planning, evidence verification, and termination decisions must be coordinated.

Off-policy distillation instead provides proactive experience-guided supervision.
Because the experience-aware teacher generates the full trajectory from the beginning, experience can shape the entire decision sequence, from initial search planning to final answering.
After rejection sampling, the student is trained on compact and successful trajectories that directly demonstrate end-to-end experience-guided behavior.
This yields a cleaner supervision signal that is better aligned with the behavior we aim to internalize.

\subsubsection{Rollout Cost and Trajectory Efficiency}

The two regimes also differ in effective rollout cost.
We control the query-level rollout budget by using the same set of rollout queries for both regimes, but the actual interaction cost largely depends on trajectory length.

\begin{table}[!t
]
\centering
\small
\setlength{\tabcolsep}{4pt}
\renewcommand{\arraystretch}{1.1}
\begin{tabular}{lccc}
\toprule
 & \textbf{Base} & \textbf{Teacher} & \makecell{\textbf{Updated}\\\textbf{Student}} \\
\midrule
Avg. assistant turns & 2.5 & 4.5 & 21.9 \\
\bottomrule
\end{tabular}
\caption{
Average assistant turns per trajectory.
The updated student is measured after one internal on-policy weight update.
}
\label{tab:assistant-turns}
\end{table}

As shown in Table~\ref{tab:assistant-turns}, after one internal weight update in on-policy context-distillation, the updated student produces substantially longer trajectories, averaging 21.9 assistant turns compared with only 2.5 for the base model and 4.5 for the experience-aware teacher.
This trajectory inflation increases the practical interaction cost of the on-policy regime, even under an identical query budget.
In contrast, off-policy context-distillation avoids this overhead by sampling shorter trajectories directly from the experience-aware teacher and applying rejection sampling to filter low-quality variants.
By leveraging concise teacher rollouts, off-policy context-distillation provides a more efficient supervision loop for iterative internalization.

\subsection{Stable Multi-Iteration Experience-Based Self-Evolution}
Having analyzed the three dimensions separately, we evaluate whether their synthesis supports stable experience-based self-evolution. Our final configuration integrates principle-level experience, step-wise injection, and off-policy context-distillation. As shown in Figure~\ref{fig:qwen_self_evolution_experience_use}, this combined design successfully sustains robust performance gains across consecutive iterations. The internalized model consistently outperforms the vanilla base model, demonstrating that experience-conditioned behavior is reliably embedded into model parameters. 

Furthermore, in-context evaluation reveals that the updated model retains its capacity to exploit the experience pool, ensuring that the student can effectively serve as the experience-aware teacher for the subsequent iteration. Unlike unstable baselines, this design simultaneously preserves standalone parametric execution and in-context responsiveness across iterative updates. Together, these three complementary dimensions form a stable recipe for multi-iteration experience internalization and sustainable self-evolution.

\section{Conclusion}
\label{sec:conclusion}

We study experience internalization beyond single-iteration transfer and show that existing methods can fail to sustain improvement across self-evolution iterations.
Through three dimensions, we find that principle-level experience provides a more durable signal than instance-level experience, step-wise injection better aligns experience with intermediate decision states, and off-policy context-distillation offers more coherent supervision than on-policy context-distillation.
Combining these findings yields a stable recipe for multi-iteration experience internalization, enabling LLM agents to better transform accumulated experience into reusable capability across self-evolution cycles.

\section*{Limitations}

Our experiments focus on web-reasoning agent tasks, so further evaluation is needed to assess whether the findings generalize to other domains, languages, and agent settings.
In addition, while we study three key dimensions of experience internalization, other factors such as experience-pool size, selector quality, and filtering criteria may also affect stability.
We leave a more comprehensive exploration of these factors to future work.

\section*{Broader Impact}

This work studies stable experience internalization for self-evolving LLM agents.
By analyzing why experience internalization can degrade across iterations, our findings may help build agents that more reliably transform accumulated experience into reusable model capability.
This can benefit long-horizon tool-use applications such as web reasoning, information seeking, and research assistance, where agents must search, verify evidence, and update their behavior from past interactions.

At the same time, more stable internalization may also reinforce undesirable behaviors if the accumulated experience contains incorrect, biased, or unsafe patterns.
This risk is especially relevant in self-evolving systems, where models repeatedly generate, internalize, and reuse their own experience.
Practical deployment should therefore include trajectory filtering, experience-pool auditing, human oversight, and restrictions in high-risk settings.
Our work focuses on improving the stability of experience internalization across self-evolution iterations, while practical deployment should still involve appropriate oversight and safeguards.

\bibliography{custom}

\appendix

\label{sec:app_exp}

\begin{figure*}[t]
    \centering
    \includegraphics[width=\linewidth]{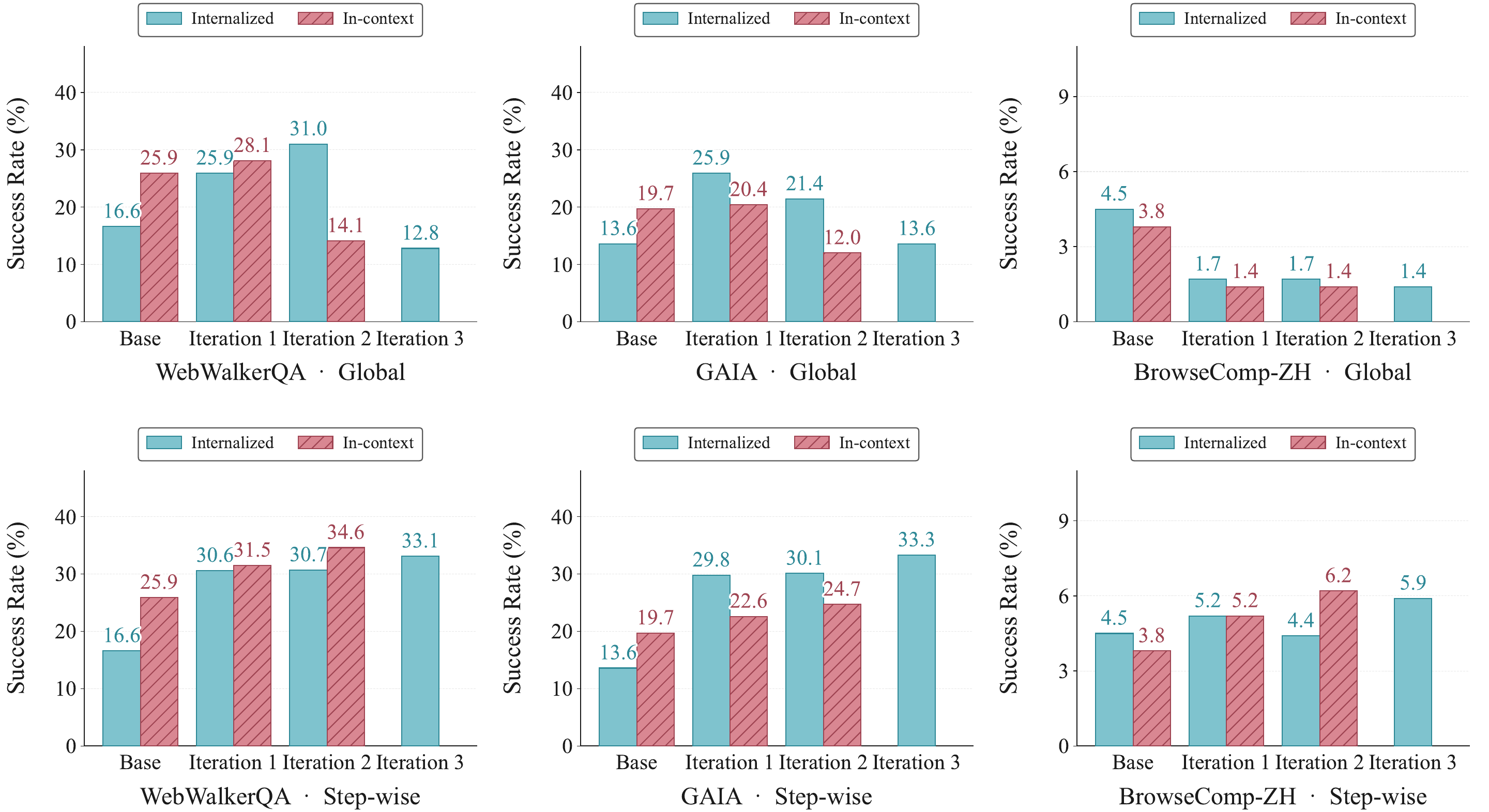}
    \caption{
    Experience internalization and in-context experience use under DeepSeek-generated principle-level experience and off-policy context-distillation.
    Top panels use global injection, and bottom panels use step-wise injection.
    Cyan bars denote internalized inference without inference-time experience, while red bars denote performance with the corresponding experience pool provided in context.
    }
    \label{fig:deepseek_experience_use}
\end{figure*}
\begin{figure*}[t]
    \centering
    \includegraphics[width=\textwidth]{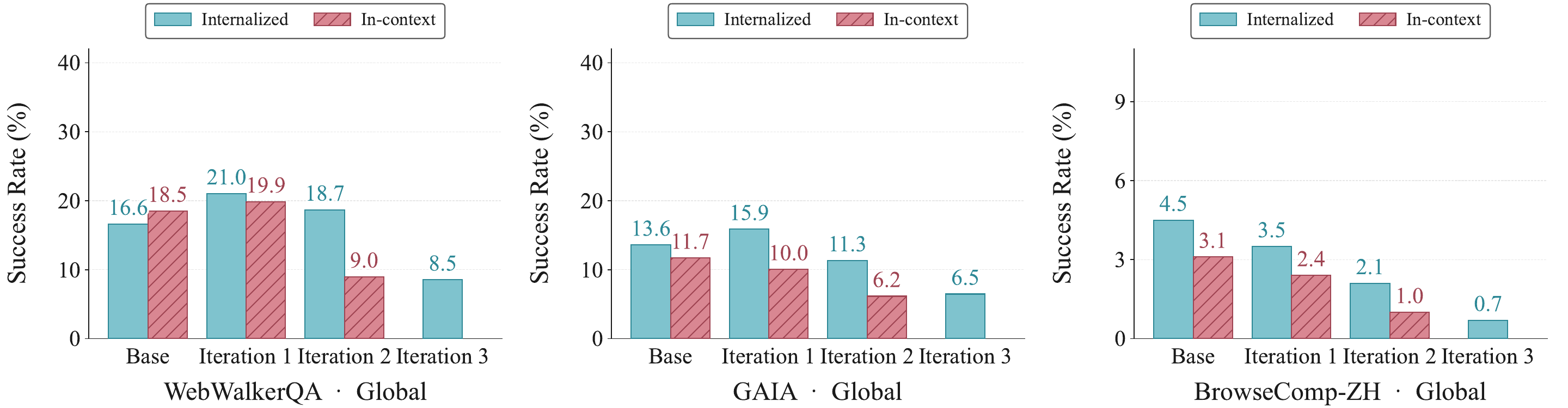}
    \caption{
    Experience internalization and in-context experience use under global injection with principle-level self-generated experience and off-policy context-distillation.
    Cyan bars denote internalized inference without inference-time experience, while red bars denote performance with the corresponding experience pool provided in context.
    }
    \label{fig:global_experience_use}
\end{figure*}
\section{Statement on the Use of LLMs}

We used LLMs in two ways in this work.
First, LLMs were used as writing assistants to polish the manuscript, improve grammar, and refine presentation.
All technical claims, experimental designs, analyses, and final writing decisions were made and verified by the authors.

Second, LLMs were used within the experimental pipeline.
Specifically, DeepSeek-V4 was used to summarize agent trajectories into natural-language experience, select relevant experience for step-wise injection, and generate experience-conditioned teacher trajectories for distillation.
In the Qwen self-generated setting, the student-side Qwen model was used instead for experience extraction and selection.
These LLM-generated artifacts constitute the experience pools and teacher supervision used in our internalization experiments.

No LLM was used to generate evaluation benchmark questions, reference answers, or reported results.
All reported metrics were obtained by running the evaluated agent models under the experimental settings described in Section~\ref{sec:exp}.
The authors take full responsibility for the content, experiments, and conclusions of the paper.

\section{Implementation Details}
\label{sec:app_impl}

\paragraph{Agent Environment and Tools.}
Our agent follows the ReAct-style interaction format described in Section~\ref{sec:formulation}.
At each step, the model produces either a tool call or a terminal answer.
We provide five tools: \textit{Search}, \textit{Visit}, \textit{Python}, \textit{Scholar}, and \textit{File Parser}.
All experiments use a maximum of $T_{\max}=100$ interaction steps and a context window of 32,768 tokens.

\paragraph{Trajectory Collection.}
Training trajectories are sampled from the 15K-example web-reasoning corpus described in Section~\ref{sec:exp}.
For on-policy context-distillation, trajectories are generated by the current student model and supervised by the experience-aware teacher.
For off-policy context-distillation, trajectories are generated by the experience-aware teacher and then filtered by rejection sampling before training.

\paragraph{Experience Extraction and Selection.}
Unless otherwise specified, DeepSeek-V4 is used to summarize trajectories into natural-language experience and select relevant experience for step-wise injection.
In the Qwen self-generated setting, the student-side Qwen model is used for experience extraction and selection.
Instance-level experience preserves trajectory-specific observations and tool-use traces, whereas principle-level experience abstracts reusable strategies, search principles, and failure patterns.

\paragraph{Distillation Training.}
All training is implemented with verl~\citep{sheng2024hybridflow}.
Students are optimized with AdamW using a learning rate of $1\times10^{-5}$, a batch size of 128, and 5 training epochs on $8\times$ NVIDIA A800 GPUs.
On-policy context-distillation uses student-induced trajectories with teacher supervision at each step, while off-policy context-distillation trains on rejection-filtered teacher-generated trajectories.

\paragraph{Self-Evolution Procedure.}
We run self-evolution for three internalization iterations.
At each iteration, the current model generates trajectories, the trajectories are summarized into an updated experience pool, and the resulting experience-conditioned behavior is distilled into the next model.
Unless otherwise stated, each iteration refreshes the experience pool using trajectories generated by the current model.

\paragraph{Inference and Evaluation.}
At inference time, models are evaluated without inference-time experience unless explicitly marked as in-context experience use.
We use temperature 0.7 for generation.
WebWalkerQA and BrowseComp-ZH are evaluated with one rollout per query and reported as Pass@1.
GAIA-Text-103 is evaluated over three independent rollouts per query and reported as average accuracy.

\section{Experience-Use Ability under Different Injection Patterns}
\label{app:experience_use}

Figures~\ref{fig:deepseek_experience_use} and~\ref{fig:global_experience_use} provide additional analysis of experience-use ability across self-evolution iterations.
We first examine the setting with DeepSeek-generated principle-level experience and off-policy context-distillation.
As shown in Figure~\ref{fig:deepseek_experience_use}, even when the experience is generated by a stronger external model, global injection shows unstable in-context experience use across iterations.
In contrast, step-wise injection maintains stronger internalized performance and better preserves the model's ability to benefit from explicit experience.
This suggests that the advantage of step-wise injection is not merely due to stronger experience quality, but also to how experience is aligned with intermediate decision states.

We then examine the more challenging self-generated setting.
Figure~\ref{fig:global_experience_use} reports results under Qwen-generated principle-level experience, global injection, and off-policy context-distillation.
In this setting, global injection degrades in both experience-free inference and in-context experience use, indicating that it does not reliably preserve the model's ability to use its updated experience pool during iterative self-evolution.
Together, these results show that state-aligned experience injection is important for preserving experience-use ability across iterations, especially when the experience pool is generated by the evolving model.

\section{Complete Self-Evolution Results}
\label{app:complete_results}

Table~\ref{tab:self-evolution-full} reports the complete self-evolution results across experience sources, injection patterns, distillation regimes, and model backbones.
The main text presents the key comparisons used to analyze experience granularity, injection pattern, and internalization regime, while this table provides the full set of internalized and in-context inference results.
Overall, the complete results are consistent with the main findings: step-wise injection is more stable than global injection across iterations, and off-policy context-distillation provides stronger multi-iteration performance than on-policy context-distillation under the same principle-level, step-wise setting.

\definecolor{groupbg}{HTML}{E6EEF7}
\definecolor{subgroupbg}{HTML}{F4F7FB}
\definecolor{gainGreen}{HTML}{1B7F3B}
\definecolor{lossRed}{HTML}{B83232}

\providecommand{\gain}[1]{\textcolor{gainGreen}{\textsubscript{+#1}}}
\providecommand{\drop}[1]{\textcolor{lossRed}{\textsubscript{-#1}}}
\providecommand{\na}{\textcolor{gray!55}{--}}
\providecommand{\grouprow}{\rowcolor{groupbg}[0pt][0pt]}
\providecommand{\subrow}{\rowcolor{subgroupbg}}

\begin{table*}[!htbp]
\centering
\small
\renewcommand{\arraystretch}{1.15}
\setlength{\tabcolsep}{4pt}
\begin{tabular}{@{}l cccccc cccccc @{}}
\toprule
\multirow{1}{*}{\textbf{Configuration}}
& \multicolumn{6}{c}{\textbf{Internalized inference}}
& \multicolumn{6}{c}{\textbf{In-context inference}} \\
\cmidrule(lr){2-7}\cmidrule(lr){8-13}
& \multicolumn{2}{c}{WebWalkerQA} 
& \multicolumn{2}{c}{GAIA}
& \multicolumn{2}{c}{BrowseComp-ZH}
& \multicolumn{2}{c}{WebWalkerQA} 
& \multicolumn{2}{c}{GAIA}
& \multicolumn{2}{c}{BrowseComp-ZH} \\
\midrule

\grouprow
\multicolumn{13}{@{}l@{}}{\textbf{Qwen3-4B-Instruct-2507} \textit{(base model)}} \\
\quad w/o experience  
& \multicolumn{2}{c}{16.6} 
& \multicolumn{2}{c}{13.6}
& \multicolumn{2}{c}{4.5}
& \multicolumn{2}{c}{\na} 
& \multicolumn{2}{c}{\na}
& \multicolumn{2}{c}{\na} \\
\quad w/ Qwen-generated experience  
& \multicolumn{2}{c}{\na} 
& \multicolumn{2}{c}{\na}
& \multicolumn{2}{c}{\na}
& \multicolumn{2}{c}{18.5} 
& \multicolumn{2}{c}{11.7}
& \multicolumn{2}{c}{3.1} \\
\quad w/ DeepSeek-generated experience  
& \multicolumn{2}{c}{\na} 
& \multicolumn{2}{c}{\na}
& \multicolumn{2}{c}{\na}
& \multicolumn{2}{c}{25.9} 
& \multicolumn{2}{c}{19.7}
& \multicolumn{2}{c}{3.8} \\
\midrule

\grouprow
\multicolumn{13}{@{}l@{}}{\textbf{Qwen3-8B-Instruct} \textit{(base model)}} \\
\quad w/o experience  
& \multicolumn{2}{c}{21.8} 
& \multicolumn{2}{c}{18.5}
& \multicolumn{2}{c}{4.5}
& \multicolumn{2}{c}{\na} 
& \multicolumn{2}{c}{\na}
& \multicolumn{2}{c}{\na} \\
\quad w/ DeepSeek-generated experience  
& \multicolumn{2}{c}{\na} 
& \multicolumn{2}{c}{\na}
& \multicolumn{2}{c}{\na}
& \multicolumn{2}{c}{27.79} 
& \multicolumn{2}{c}{26.21}
& \multicolumn{2}{c}{4.2} \\
\midrule

\grouprow
\multicolumn{13}{@{}l@{}}{\textbf{Qwen-generated experience} $\,\bullet\,$ \textit{Global injection}} \\
\subrow
\multicolumn{13}{l}{\textit{Qwen3-4B-Instruct-2507} $\,\bullet\,$ \textit{Off-policy distillation}} \\
\quad iter 1 
& \multicolumn{2}{c}{21.0} 
& \multicolumn{2}{c}{15.9}
& \multicolumn{2}{c}{3.5}
& \multicolumn{2}{c}{19.9} 
& \multicolumn{2}{c}{10.0}
& \multicolumn{2}{c}{2.4} \\
\quad iter 2 
& \multicolumn{2}{c}{18.7} 
& \multicolumn{2}{c}{11.3}
& \multicolumn{2}{c}{2.1}
& \multicolumn{2}{c}{9.0}  
& \multicolumn{2}{c}{6.2}
& \multicolumn{2}{c}{1.0} \\
\quad iter 3 
& \multicolumn{2}{c}{8.5}  
& \multicolumn{2}{c}{6.5}
& \multicolumn{2}{c}{0.7}
& \multicolumn{2}{c}{\na}   
& \multicolumn{2}{c}{\na}
& \multicolumn{2}{c}{\na} \\
\midrule

\grouprow
\multicolumn{13}{@{}l@{}}{\textbf{Qwen-generated experience} $\,\bullet\,$ \textit{Step-wise injection}} \\
\subrow
\multicolumn{13}{l}{\textit{Qwen3-4B-Instruct-2507} $\,\bullet\,$ \textit{Off-policy distillation}} \\
\quad iter 1 
& \multicolumn{2}{c}{29.0} 
& \multicolumn{2}{c}{22.7}
& \multicolumn{2}{c}{5.2}
& \multicolumn{2}{c}{32.1} 
& \multicolumn{2}{c}{24.9}
& \multicolumn{2}{c}{5.5} \\
\quad iter 2 
& \multicolumn{2}{c}{28.5} 
& \multicolumn{2}{c}{24.0}
& \multicolumn{2}{c}{4.5}
& \multicolumn{2}{c}{31.3} 
& \multicolumn{2}{c}{23.3}
& \multicolumn{2}{c}{5.2} \\
\quad iter 3 
& \multicolumn{2}{c}{30.0} 
& \multicolumn{2}{c}{24.6}
& \multicolumn{2}{c}{5.9}
& \multicolumn{2}{c}{\na}   
& \multicolumn{2}{c}{\na}
& \multicolumn{2}{c}{\na} \\
\midrule

\grouprow
\multicolumn{13}{@{}l@{}}{\textbf{DeepSeek-generated experience} $\,\bullet\,$ \textit{Global injection}} \\
\subrow
\multicolumn{13}{l}{\textit{Qwen3-4B-Instruct-2507} $\,\bullet\,$ \textit{Off-policy distillation}} \\
\quad iter 1 
& \multicolumn{2}{c}{25.9} 
& \multicolumn{2}{c}{25.9}
& \multicolumn{2}{c}{1.7}
& \multicolumn{2}{c}{28.1} 
& \multicolumn{2}{c}{20.4}
& \multicolumn{2}{c}{1.4} \\
\quad iter 2 
& \multicolumn{2}{c}{31.0} 
& \multicolumn{2}{c}{21.4}
& \multicolumn{2}{c}{1.7}
& \multicolumn{2}{c}{14.1} 
& \multicolumn{2}{c}{12.0}
& \multicolumn{2}{c}{1.4} \\
\quad iter 3 
& \multicolumn{2}{c}{12.8} 
& \multicolumn{2}{c}{13.6}
& \multicolumn{2}{c}{1.4}
& \multicolumn{2}{c}{\na}   
& \multicolumn{2}{c}{\na}
& \multicolumn{2}{c}{\na} \\

\subrow
\multicolumn{13}{l}{\textit{Qwen3-4B-Instruct-2507} $\,\bullet\,$ \textit{On-policy distillation}} \\
\quad iter 1 
& \multicolumn{2}{c}{29.0} 
& \multicolumn{2}{c}{22.8}
& \multicolumn{2}{c}{3.1}
& \multicolumn{2}{c}{\na} 
& \multicolumn{2}{c}{\na}
& \multicolumn{2}{c}{\na} \\
\quad iter 2 
& \multicolumn{2}{c}{22.5} 
& \multicolumn{2}{c}{19.4}
& \multicolumn{2}{c}{3.8}
& \multicolumn{2}{c}{\na} 
& \multicolumn{2}{c}{\na}
& \multicolumn{2}{c}{\na} \\
\quad iter 3 
& \multicolumn{2}{c}{19.9} 
& \multicolumn{2}{c}{18.1}
& \multicolumn{2}{c}{3.5}
& \multicolumn{2}{c}{\na} 
& \multicolumn{2}{c}{\na}
& \multicolumn{2}{c}{\na} \\
\midrule

\grouprow
\multicolumn{13}{@{}l@{}}{\textbf{DeepSeek-generated experience} $\,\bullet\,$ \textit{Step-wise injection}} \\

\subrow
\multicolumn{13}{l}{\textit{Qwen3-4B-Instruct-2507} $\,\bullet\,$ \textit{Off-policy distillation}} \\
\quad iter 1 
& \multicolumn{2}{c}{30.6} 
& \multicolumn{2}{c}{29.8}
& \multicolumn{2}{c}{5.2}
& \multicolumn{2}{c}{31.5} 
& \multicolumn{2}{c}{22.6}
& \multicolumn{2}{c}{5.2} \\
\quad iter 2 
& \multicolumn{2}{c}{30.7} 
& \multicolumn{2}{c}{30.1}
& \multicolumn{2}{c}{4.4}
& \multicolumn{2}{c}{34.6} 
& \multicolumn{2}{c}{24.7}
& \multicolumn{2}{c}{6.2} \\
\quad iter 3 
& \multicolumn{2}{c}{33.1} 
& \multicolumn{2}{c}{33.3}
& \multicolumn{2}{c}{5.9}
& \multicolumn{2}{c}{\na}   
& \multicolumn{2}{c}{\na}
& \multicolumn{2}{c}{\na} \\

\subrow
\multicolumn{13}{l}{\textit{Qwen3-4B-Instruct-2507} $\,\bullet\,$ \textit{On-policy distillation}} \\
\quad iter 1 
& \multicolumn{2}{c}{35.0} 
& \multicolumn{2}{c}{28.8}
& \multicolumn{2}{c}{3.8}
& \multicolumn{2}{c}{\na} 
& \multicolumn{2}{c}{\na}
& \multicolumn{2}{c}{\na} \\
\quad iter 2 
& \multicolumn{2}{c}{32.4} 
& \multicolumn{2}{c}{27.2}
& \multicolumn{2}{c}{6.6}
& \multicolumn{2}{c}{\na} 
& \multicolumn{2}{c}{\na}
& \multicolumn{2}{c}{\na} \\
\quad iter 3 
& \multicolumn{2}{c}{31.5} 
& \multicolumn{2}{c}{25.6}
& \multicolumn{2}{c}{3.8}
& \multicolumn{2}{c}{\na} 
& \multicolumn{2}{c}{\na}
& \multicolumn{2}{c}{\na} \\

\subrow
\multicolumn{13}{l}{\textit{Qwen3-8B-Instruct} $\,\bullet\,$ \textit{Off-policy distillation}} \\
\quad iter 1 
& \multicolumn{2}{c}{32.9} 
& \multicolumn{2}{c}{30.1}
& \multicolumn{2}{c}{4.8}
& \multicolumn{2}{c}{\na} 
& \multicolumn{2}{c}{\na}
& \multicolumn{2}{c}{\na} \\
\quad iter 2 
& \multicolumn{2}{c}{31.8} 
& \multicolumn{2}{c}{28.8}
& \multicolumn{2}{c}{4.2}
& \multicolumn{2}{c}{\na} 
& \multicolumn{2}{c}{\na}
& \multicolumn{2}{c}{\na} \\
\quad iter 3 
& \multicolumn{2}{c}{34.6} 
& \multicolumn{2}{c}{29.8}
& \multicolumn{2}{c}{6.6}
& \multicolumn{2}{c}{\na} 
& \multicolumn{2}{c}{\na}
& \multicolumn{2}{c}{\na} \\

\subrow
\multicolumn{13}{l}{\textit{Qwen3-8B-Instruct} $\,\bullet\,$ \textit{On-policy distillation}} \\
\quad iter 1 
& \multicolumn{2}{c}{33.9} 
& \multicolumn{2}{c}{28.5}
& \multicolumn{2}{c}{4.5}
& \multicolumn{2}{c}{\na} 
& \multicolumn{2}{c}{\na}
& \multicolumn{2}{c}{\na} \\
\quad iter 2 
& \multicolumn{2}{c}{31.5} 
& \multicolumn{2}{c}{27.8}
& \multicolumn{2}{c}{1.4}
& \multicolumn{2}{c}{\na} 
& \multicolumn{2}{c}{\na}
& \multicolumn{2}{c}{\na} \\
\quad iter 3 
& \multicolumn{2}{c}{32.2} 
& \multicolumn{2}{c}{23.9}
& \multicolumn{2}{c}{1.4}
& \multicolumn{2}{c}{\na} 
& \multicolumn{2}{c}{\na}
& \multicolumn{2}{c}{\na} \\
\bottomrule
\end{tabular}
\caption{
Self-evolution results under different experience sources, injection patterns, and distillation regimes.
}

\label{tab:self-evolution-full}
\end{table*}
\end{document}